# Improving the Accuracy and Efficiency of MAP Inference for Markov Logic

**Sebastian Riedel**
Institute for Collaborating and Communicating Systems
School of Informatics
University of Edinburgh
EH81BL Edinburgh, Scotland

## Abstract

In this work we present Cutting Plane Inference (CPI), a Maximum A Posteriori (MAP) inference method for Statistical Relational Learning. Framed in terms of Markov Logic and inspired by the Cutting Plane Method, it can be seen as a meta algorithm that instantiates small parts of a large and complex Markov Network and then solves these using a conventional MAP method. We evaluate CPI on two tasks, Semantic Role Labelling and Joint Entity Resolution, while plugging in two different MAP inference methods: the current method of choice for MAP inference in Markov Logic, MaxWalkSAT, and Integer Linear Programming. We observe that when used with CPI both methods are significantly faster than when used alone. In addition, CPI improves the accuracy of MaxWalkSAT and maintains the exactness of Integer Linear Programming.

## 1 INTRODUCTION

Many tasks in Machine Learning are inherently relational: the label given to an object often depends on labels given to a set of related objects. For example, in Semantic Role Labelling [Carreras and Marquez, 2005] we are asked to label phrases with the role they play with respect to a given verb. Here the role given to one phrase depends on roles we have assigned to other phrases in the same sentence. It is, for instance, not possible to have two phrases both labelled as the agent of the same verb.

Statistical Relational Learning [SRL, Ng and Subrahmanian, 1992, Koller, 1999] seeks to provide generic, solid and efficient means to solve such relational tasks. It typically uses variants of First Order Logic to describe Graphical Models with repetitive structure in a compact fashion. This has two main benefits.

Firstly, the meta-information provided by the first order model can be used to avoid a full instantiation of the Graphical Model in inference and learning. This can yield faster runtime and higher accuracy [Koller, 1999, de Salvo Braz et al., 2005, Singla and Domingos, 2006b]. Secondly, an SRL language along with a powerful interpreter allows application developers to focus on models, and machine learning researchers to focus on foundations. This paradigm of decoupling applications and algorithms has increased the speed of development in many domains [Domingos, 2006].

Markov Logic [ML, Richardson and Domingos, 2005] is an expressive SRL language that combines First Order Logic and Markov Networks. It has been successfully used for several tasks such as Information Extraction [Poon and Domingos, 2007] and Entity Resolution [Singla and Domingos, 2006a].

For most Markov Logic applications we need to solve the Maximum A Posteriori (MAP) problem of finding the most likely solution given some observation. Richardson and Domingos [2005] proposed the use of MaxWalkSAT [MWS, Kautz et al., 1996] to tackle this problem. In our experiments we apply MWS to two rather simple ML models, one for Semantic Role Labelling and one for Joint Entity Resolution. Here we found MWS to be both slow and inaccurate. However, before languages like ML can ever be used to solve tasks like joint inference in large Natural Language Processing applications [Domingos, 2007] we surely need to able to efficiently and accurately solve comparatively simple ones.

Rather than investigating the use of other solvers such as Belief Propagation and its variants, which can perform quite poorly for the large, densely connected and partially deterministic networks we encounter, we focus on tackling this problem by introducing a *meta* algorithm: *Cutting Plane Inference* (CPI) inspired by the Cutting Plane Method [Dantzig et al., 1954].

CPI incrementally instantiates only those portions of the complete Markov Network for which a current solution can be further optimised and solves these using an existing inference method. Often these partial problems are significantly smaller and less complex. Consequently, they are more easily solved than the complete problem.

Empirically we show that for Semantic Role Labelling CPI plus MWS is significantly faster and more accurate than MWS alone. When used with Integer Linear Programming (ILP), CPI achieves optimal accuracy due to the exactness of ILP, yet runs significantly faster than when using ILP alone. With this accuracy we are able to achieve state-of-the-art results in Semantic Role Labelling with minimal engineering effort. When tested on an Joint Entity Resolution model taken from from the Markov Logic literature [Singla and Domingos, 2005] CPI with MWS does again better than MWS alone both in terms of speed and accuracy. CPI with ILP allows us to perform efficient and *exact* inference on this task while ILP alone is infeasible.

In the next section of this paper we will present Markov Logic. Section 3 shows two ways of solving the MAP problem for the Markov Networks that Markov Logic describes: MWS and ILP. In section 4 Cutting Plane Inference is presented and we formally show how the accuracy of CPI depends on the accuracy of the base solver. Section 5 compares CPI in combination with MWS and ILP to plain MWS and ILP on two tasks. The first is Semantic Role Labelling, the second Joint Entity Resolution. We conclude with section 6.

## 2   MARKOV LOGIC

Markov Logic [ML, Richardson and Domingos, 2005] is an SRL language based on First Order Logic and Markov Networks. It can be seen as a formalism that extends First Order Logic to allow formulae that can be violated with some penalty. From an alternative point of view, it is a expressive template language that uses First Order Logic formulae to instantiate Markov Networks of repetitive structure. [1]

Let us introduce some notation by example. Assume a simplified version of Semantic Role Labelling where we use an unary predicate *agent* to select the constituent that acts as agent for a given verb. We also maintain a set of additional predicates that provide information about constituents and their relation to the verb. For example, *left* can be a unary predicate that denotes constituents to the left of the verb. The set of all predicates will be called $P$. We also maintain a *finite* set $C$ of constants representing constituents, words, tags etc.

In the following we will use $n_p$ to denote the arity of a predicate $p$, and thus $n_{left} = n_{agent} = 1$. In formulae we will denote logical variables using the letter $v$ and some subscript such as $v_1$. For example, in

$$\phi_1 : agent(v_1) \Rightarrow left(v_1)$$

$v_1$ is a variable and in

$$\phi_2 : v_1 \neq v_2 \wedge agent(v_1) \Rightarrow \neg agent(v_2)$$

$v_1$ and $v_2$ are variables. The number of free variables of a formula $\phi$ will be denoted with $n_\phi$, thus $n_{\phi_1} = 1$ and $n_{\phi_2} = 2$. A *grounding* $\phi\left[v_1/c_1, \ldots, v_{n_\phi}/c_{n_\phi}\right]$ is generated by replacing each occurrence of each $v_i$ with the constant $c_i$. We will often write $\phi[\mathbf{v}/\mathbf{c}]$ to mean $\phi\left[v_1/c_1, \ldots, v_{n_\phi}/c_{n_\phi}\right]$. For example, $\phi_2[\mathbf{v}/\mathbf{c}] = c_1 \neq c_2 \wedge agent(c_1) \Rightarrow \neg agent(c_2)$.

A formula that does not contain any variables is *ground*. A formula that contains a single predicate and nothing else is an *atom*. A set of ground atoms is called a *possible world* [Genesereth and Nilsson, 1987]. We say that a possible world $W$ *satisfies a formula* $\phi$ and write $\vDash_W \phi$ if $\phi$ is true in $W$. For example, the possible world $\{agent(c_1)\}$ does not satisfy $\phi_1[v_1/c_1]$; the possible world $\{left(c_1), agent(c_1)\}$ does. In the following we will identify the binary vector $\mathbf{y} = \left(y_{p(\mathbf{c})}\right)_{p \in P, \mathbf{c} \in C^{n_p}}$ with the possible world $\{p(\mathbf{c}) | y_{p(\mathbf{c})} = 1\}$. For the sake of brevity we will often write $y_\mathbf{c}^p$ instead of $y_{p(\mathbf{c})}$. The set of all possible worlds we can construct using a set of predicates $P$ and a set of constants $C$ is $\mathcal{Y}_{P,C}$.

In First Order Logic a knowledge base is a set of formulae. It describes the set of possible worlds that for which all its formulae are satisfied. In Markov Logic the equivalent of a first order knowledge base is a *Markov Logic Network* (MLN). Instead of classifying models as either consistent (all formulae are satisfied) or inconsistent (some are not) an MLN maps each possible world to a probability. This allows us to model uncertainty in our beliefs about the world. For example, a world in which $agent(c_1) \Rightarrow left(c_1)$ does not hold should not be impossible, it should just be a bit less likely because the agent of a verb tends to be on its left but can appear on its right in passive constructions.

We define an MLN $M$ as set of pairs $\{(\phi_i, w_i)\}_i$ where each $\phi_i$ is a formula in First Order Logic and $w_i \in \mathbb{R}$ is a real number. Together with a *finite* set of constants $C$, an MLN $M$ defines a log-linear probability distribution over possible worlds $\mathbf{y} \in \mathcal{Y}_{P,C}$ as follows

$$p(\mathbf{y}) = \frac{1}{Z} \exp \left( \sum_{(\phi,w) \in M} w \sum_{\mathbf{c} \in C^{n_\phi}} f_\mathbf{c}^\phi (\mathbf{y}) \right) \quad (1)$$

where the feature function $f_\mathbf{c}^\phi$ is defined as

$$f_\mathbf{c}^\phi(\mathbf{y}) = \mathbb{I}\left(\vDash_\mathbf{y} \phi\left[v_1/c_1, \ldots, v_{n_\phi}/c_{n_\phi}\right]\right)$$

---

[1] Note that while this paper focuses on Markov Logic due to its expressive power and possibility of undirected dependencies, we note that much of the work reported here can be transferred to other formalisms.

$Z$ is a normalisation constant, $\mathbb{I}(true) = 1$ and $\mathbb{I}(false) = 0$.

This distribution is strictly positive and corresponds to a Markov Network which is referred to as the *Ground Markov Network*. Note that we can represent hard constraints using very large weights.

For example, with $M = \{(\phi_1, 2.5), (\phi_2, 1.2)\}$ and the finite set of constants $C = \{n_1, n_2, \ldots\}$ that represent the nodes of the parse tree, the log-linear model would contain, among others, the feature $f_{n_1}^{\phi_1}(\mathbf{y}) = \mathbb{I}(\vDash_{\mathbf{y}} agent(n_1) \Rightarrow left(n_1))$ that returns 1 if the contained ground formula holds in the possible world $\mathbf{y}$ and 0 otherwise.

## 3 MAP INFERENCE

In many settings we are given an MLN $M$ and the state of a set of *observed* ground atoms $(x_{p(\mathbf{c})})_{p \in O, \mathbf{c} \in C^{n_p}}$ for a set of observable predicates $O$. We are then asked to find the set of *hidden* ground atoms $\hat{\mathbf{y}} \in \mathcal{Y}_{H,C}$ for a set of remaining predicates $H = P \setminus O$ with maximum *a posteriori* probability (MAP)[2]

$$\hat{\mathbf{y}} = \arg\max_{\mathbf{y} \in \mathcal{Y}_{H,C}} p(\mathbf{y}|\mathbf{x}) = \arg\max_{\mathbf{y} \in \mathcal{Y}_{H,C}} s(\mathbf{y}, \mathbf{x})$$

where

$$s(\mathbf{y}, \mathbf{x}) = \sum_{(\phi, w) \in M} w \sum_{\mathbf{c} \in C^{n_\phi}} f_{\mathbf{c}}^{\phi}(\mathbf{y}, \mathbf{x}) \quad (2)$$

can be considered as a linear *discriminant* or *scoring* function that evaluates the goodness of a problem solution pair $(\mathbf{x}, \mathbf{y})$.

For example, we might be looking for the truth states of the $H = \{agent\}$ atoms while knowing the state of all $O = \{left\}$ ground atoms that indicate which constituents are placed to the left of the verb.

### 3.1 MAXWALKSAT

Previous work [Richardson and Domingos, 2005] finds $\hat{\mathbf{y}}$ using MaxWalkSAT (MWS), an approximate Random Walk method that has been very successfully used to solve Weighted SAT Problems [Kautz et al., 1996].

It starts by assigning a random state to all ground atoms and proceeds by repeatedly picking a random unsatisfied ground clause. With probability $q$ a random ground atom of this clause is picked and its state is flipped. With probability $1 - q$ the ground atom which, when flipped, causes the largest increase in total weight $s(\mathbf{y}, \mathbf{x})$ is chosen to be flipped. The process is repeated until a fixed number $n_{flips}$ of flips is reached. Optionally one can try again $n_{restarts}$ times to find a better $\mathbf{y}$, each time starting at a new random solution.

---

[2] In the case where multiple maxima exist we can pick any of these.

### 3.2 INTEGER LINEAR PROGRAMMING

Integer Linear Programming [ILP, Winston and Venkataramanan, 2003] refers to the process of optimising a linear objective function under a set of linear inequalities and the constraint that all (or some) variables are integers. ILP has been used in many tasks to solve MAP problems [Roth and Yih, 2005, Clarke and Lapata, 2007] because of its exactness, its declarative nature and the availability of very efficient ILP solvers.

Here we present a generic mapping from Ground Markov Networks in Markov Logic to ILPs.[3] We start by replacing each feature function application $f_{\mathbf{c}}^{\phi}(\mathbf{y})$ in equation 2 with a binary variable $\lambda_{\mathbf{c}}^{\phi}$ and constrain $f_{\mathbf{c}}^{\phi}(\mathbf{y})$ and $\lambda_{\mathbf{c}}^{\phi}$ to be equal. This leads to the optimisation problem

$$\arg\max_{\mathbf{y} \in \mathcal{Y}_{H,C}} \sum_{(\phi, w) \in M} w \sum_{\mathbf{c} \in C^{n_\phi}} \lambda_{\mathbf{c}}^{\phi}$$
$$s.t \quad \lambda_{\mathbf{c}}^{\phi} = f_{\mathbf{c}}^{\phi}(\mathbf{y}, \mathbf{x}) \, \forall (\phi, w) \in M, \mathbf{c} \in C^{n_\phi}$$

with a linear objective function under a set of constraints.

In order to turn this into an ILP we need to transform each constraint into a set of linear constraints over $\mathbf{y}$ and the auxiliary variables $(\lambda_{\mathbf{c}}^{\phi})_{\phi, \mathbf{c} \in C^{n_\phi}}$. This can be achieved by

1. Mapping each constraint to a logical equivalence of auxiliary variable and ground formula, such as $\lambda_{n_1}^{\phi} \Leftrightarrow agent(n_1) \Rightarrow left(n_1)$

2. Replacing ground atoms by either *true* or *false* if they are observed or, if not, by their corresponding decision variable, as in
$\lambda_{n_1}^{\phi} \Leftrightarrow y_{n_1}^{agent} \Rightarrow false$

3. Transforming the logical equivalence into Conjunctive Normal Form[4], as in $\left(\neg\lambda_{n_1}^{\phi} \vee y_{n_1}^{agent}\right) \wedge \left(\lambda_{n_1}^{\phi} \vee \neg y_{n_1}^{agent}\right)$

4. Replacing each disjunction by a linear constraint [Williams, 1999], for example
$-1.0 \cdot \lambda_{n_1}^{\phi} + 1.0 \cdot y_{n_1}^{agent} \geq 0$

Note that we can significantly simplify the above program for hard constraints (i.e.., formulae with very large $w$) and formulae with only one hidden atom. We omit details for brevity.

---

[3] In general it is always possible to map any Markov Network to an ILP [Taskar, 2004]. However, our mapping is tailor-made for Markov Logic and yields smaller programs.

[4] If the formula contains universal (existential) quantified formulae we replace these with conjunctions (disjunctions) using the finite set of constants $C$.

# 4 CUTTING PLANE INFERENCE

We will show in the empirical section of this paper that running inference using the full grounding of a Markov Logic Network can be slow and in the case of MWS also inaccurate. We will now present an algorithm that tries to overcome this problem by instantiating only portions of the complete Ground Markov Network and running an off-the-shelf inference method in this network.

## 4.1 ALGORITHM

The proposed algorithm is a variant of the Cutting Plane approach from Operations Research [Dantzig et al., 1954]. Cutting Plane algorithms solve large scale constrained optimisation problems by only considering a subset of constraints. In each iteration the solution to a partial problem is provided to an oracle that returns a set of constraints[5] the solution violates. The current problem is extended by these new constraints and re-solved. The process is repeated until no more violated constraints can be found.

Instead of searching for violated constraints, *Cutting Plane Inference* (CPI) searches for feature-weight products in equation 2 that do not maximally contribute to the overall sum given the current solution. More precisely, for each formula $\phi$ and a given $(\mathbf{y}', \mathbf{x})$ we are looking for all tuples, $\text{Separate}(\phi, w, \mathbf{y}, \mathbf{x}) \subseteq C^{n_\phi}$, for which

$$w \cdot f_{\mathbf{c}}^{\phi}(\mathbf{y}', \mathbf{x}) < \max_{\mathbf{y} \in Y_{H,C}} w \cdot f_{\mathbf{c}}^{\phi}(\mathbf{y}, \mathbf{x}) \qquad (3)$$

We will say that the corresponding ground formulae are *not maximally satisfied* in the world $\mathbf{y}'$.

In the terminology of the Cutting Plane method this step is often referred to as *separation*: it finds a set of constraints that separates feasible solution from infeasible solutions. In our case this step will help to separate possible worlds with high score from those with low score.

It will be useful to define a *partial grounding* $\mathbf{G} = (G_\phi)_{(\phi,w) \in M}$ with $G_\phi \subseteq C^{n_\phi}$ that maps each first order formula $\phi$ to a set of tuples we ground it with. A partial grounding induces a *partial score*

$$s_{\mathbf{G}}(\mathbf{y}, \mathbf{x}) = \sum_{(\phi,w) \in M} w \sum_{\mathbf{c} \in G_\phi} f_{\mathbf{c}}^{\phi}(\mathbf{y}, \mathbf{x}) \qquad (4)$$

CPI proceeds as described in algorithm 1. In each iteration $i$ we maintain a partial grounding $\mathbf{G}^i$. Initially $\mathbf{G}^0$ is filled with a small number of groundings. A natural choice are all groundings of formulae that only

---
[5]In case of linear constraints these constraints form hyperplanes that further *cut* the space of feasible solution, hence the name.

---

contain one hidden predicate. In this case maximising $s_{\mathbf{G}^0}$ is trivial because the hidden variables do not interact and often gives a very good first guess.

In step 5 we find a solution $\mathbf{y}$ that maximises the partial score $s_{\mathbf{G}^{i-1}}$ (or approximately maximises it). For this we can pick our optimisation method of choice. In steps 9 and 10 we find the ground formulae which are not maximally satisfied in the current solution $\mathbf{y}$ and add them to the current partial grounding. We terminate if no more new ground formulae are found or a maximum number of iterations is reached. This process calculates one solution $\mathbf{y}$ in each iteration. The final result is the solution $\mathbf{y}$ with highest score.

---
**Algorithm 1** CPI$(M, \mathbf{G}^0, \mathbf{x})$
---
1: $i \leftarrow 0$
2: $\mathbf{y}' \leftarrow \mathbf{0}$
3: **repeat**
4: $\quad i \leftarrow i + 1$
5: $\quad \mathbf{y} \leftarrow \text{solve}\left(\mathbf{G}^{i-1}, \mathbf{x}\right)$
6: $\quad$ **if** $s(\mathbf{y}, \mathbf{x}) > s(\mathbf{y}', \mathbf{x})$ **then**
7: $\quad\quad \mathbf{y}' \leftarrow \mathbf{y}$
8: $\quad$ **end if**
9: $\quad$ **for** each $(\phi, w) \in M$ **do**
10: $\quad\quad \mathbf{G}_\phi^i \leftarrow \mathbf{G}_\phi^{i-1} \cup \text{Separate}(\phi, w, \mathbf{y}, \mathbf{x})$
11: $\quad$ **end for**
12: **until** $\mathbf{G}_\phi^i = \mathbf{G}_\phi^{i-1}$ or $i > maxIterations$
13: **return** $\mathbf{y}'$

---

The following theorem shows that when CPI returns the solution of iteration $i$ the error is bound by the sum of the error of the base solver on the partial problem and the sum of absolute weights of newly found ground formulae at this iteration. In particular, for an iteration with no more newly found groundings the error is only bound by (in fact it is equal to) the error of the base solver on the partial problem, which is likely to be much smaller and easier to solver than the original one.

This also shows that if the base solver is exact (like ILP) and no more groundings are found, CPI will be exact. If we choose a solution for which new ground formulae were found the error bound is incremented by the sum of the absolute weights of these ground clauses. Thus we still do well if the remaining clauses have small weight.

**Theorem.** *Let $\hat{\mathbf{y}}$ be an optimal solution, $\mathbf{y}'$ the solution returned by CPI taken from iteration $i$, $\hat{\mathbf{y}}_{\mathbf{G}^{i-1}}$ an optimal solution for $s_{\mathbf{G}^{i-1}}$ and $b = \sum_{(\phi,w)} \sum_{\mathbf{c} \in \mathbf{G}_\phi^i \setminus \mathbf{G}_\phi^{i-1}} |w|$ then*

$$s(\hat{\mathbf{y}}, \mathbf{x}) - s(\mathbf{y}', \mathbf{x})$$
$$\leq s_{\mathbf{G}^{i-1}}(\hat{\mathbf{y}}_{\mathbf{G}^{i-1}}, \mathbf{x}) - s_{\mathbf{G}^{i-1}}(\mathbf{y}', \mathbf{x}) + b$$

*Proof.* Let $\mathbf{G}^i \setminus \mathbf{G}^{i-1} = \left(G_\phi^i \setminus G_\phi^{i-1}\right)_\phi$ be the newly

added groundings and $\overline{\mathbf{G}^i} = \left(C^{n_\phi} \setminus G^i_\phi\right)_\phi$ the remaining groundings. We can split $s(\hat{\mathbf{y}}, \mathbf{x}) - s(\mathbf{y}', \mathbf{x})$ into three parts, a score difference for the ground formulae in $\mathbf{G}^{i-1}$, those in $\mathbf{G}^i \setminus \mathbf{G}^{i-1}$ and $\overline{\mathbf{G}^i}$. We know that $\mathbf{y}'$ solves $s_{\overline{\mathbf{G}^i}}$ optimally based on equation 3, thus $s_{\overline{\mathbf{G}^i}}(\hat{\mathbf{y}}, \mathbf{x}) - s_{\overline{\mathbf{G}^i}}(\mathbf{y}', \mathbf{x}) \leq 0$. Furthermore, in the worst case each term $w \cdot f^\phi_\mathbf{c}(\mathbf{y}', \mathbf{x})$ in $s_{\mathbf{G}^i/\mathbf{G}^{i-1}}(\mathbf{y}', \mathbf{x})$ is by $|w|$ smaller than each corresponding term in $s_{\mathbf{G}^i \setminus \mathbf{G}^{i-1}}(\hat{\mathbf{y}}, \mathbf{x})$, leading to $s_{\mathbf{G}^i \setminus \mathbf{G}^{i-1}}(\hat{\mathbf{y}}, \mathbf{x}) - s_{\mathbf{G}^i \setminus \mathbf{G}^{i-1}}(\mathbf{y}', \mathbf{x}) \leq \sum_{(\phi, w)} \sum_{\mathbf{c} \in G^i_\phi \setminus G^{i-1}_\phi} |w|$. □

Note that we do not make any claims about the runtime of the algorithm. Even without a limit on the number iterations it is guaranteed to converge in a finite number of steps due to the fact that the solution space is finite and we will either try each solution or return to a previous one. However, we cannot provide guarantees as to how many steps this will take. Thus we allow the algorithm to terminate before convergence is reached.

### 4.2 SEPARATION

An integral part of CPI is the separation step, in which we need to find all groundings $\mathbf{c}$ of a formula $\phi$ and weight $w$ which are not maximally satisfied (according to equation 3) for a given solution $\mathbf{y}'$. It is this step for which the Statistical Relational Learning paradigm comes into play. In a (propositional) Markov Network we do not have any higher order descriptions of its features. Performing separation then means evaluating all features of the network.

In Markov Logic, however, we can do better. There are two cases to consider. If $w > 0$ we have to find assignments $\mathbf{c}$ with $f^\phi_\mathbf{c}(\mathbf{y}, \mathbf{x}) = 0$, that is, groundings for which $\vDash_{\mathbf{y}, \mathbf{x}} \phi[\mathbf{v}/\mathbf{c}]$ is false. Correspondingly, for $w < 0$ we have to find $\mathbf{c}$ for which $\vDash_{\mathbf{y}, \mathbf{x}} \phi[\mathbf{v}/\mathbf{c}]$ is true.

We cast this into a database query evaluation problem[6] and store the atoms in $\mathbf{y}$ and $\mathbf{x}$ as rows in database tables. Then we convert the formula $\phi$ (or $\neg \phi$) to a database query which is executed during CPI. Such queries can often be processed very efficiently [Grohe et al., 2001]. In our experiments the cost of query evaluation was marginal when compared to the cost of numeric optimisation.

### 4.3 RELATED WORK

The idea of Cutting Planes have been used in at least two ways. We can either use it to tackle ILP problems by solving their LP relaxation and, in case the solution is fractional, generate additional constraints the integer solution is known to fulfil. Or we use it to solve problems with a large number of constraints, such as ILP formulations of the Travelling Salesman Problem [Dantzig et al., 1954], without having to include all of them.

---

[6] Alternatively we could frame this problem as an instance of theorem proving, but all axioms are ground atoms and we are looking for *all* groundings for which the formula holds – Database technology is optimised for this setting.

Our work follows previous research in MAP inference [Riedel and Clarke, 2006, Anguelov et al., 2004, Tromble and Eisner, 2006, Sontag and Jaakkola, 2007] that uses Cutting Planes to avoid including all constraints in advance. However, in this work we frame, implement and evaluate the approach more generally as a meta algorithm for MAP inference in Markov Logic Networks into which we can plug-in any existing propositional solver. This includes the introduction of a separation routine that does not require additional implementation efforts when applied to new tasks. Markov Logic Networks may also contain nondeterministic constraints. In contrast to previous work [Tromble and Eisner, 2006] our method handles these without the need to branch-and-bound.

CPI is also similar in nature to LazySAT [Singla and Domingos, 2006b], a memory-efficient implementation of MWS: both methods avoid to instantiate the full ground network. However, while CPI only instantiates new parts of the ground network once the base solver has optimised the current partial network, LazySAT instantiates new parts of the network whenever they may be needed during the inner loop of MWS. Note that although CPI also reduces memory overhead, in this work we focus on its speed and accuracy and thus do not directly compare it to LazySAT, which inherits the speed and accuracy of MWS.

## 5 EXPERIMENTS

We use two tasks to evaluate the utility of CPI as a meta MAP inference method for Markov Logic. The first is Semantic Role Labelling, the second Joint Entity Resolution. In both cases we want to investigate how CPI affects the runtime and accuracy of two base solvers: MWS and ILP. For all experiments we use our own Markov Logic implementation running on a Pentium 4 at 2.8Ghz with 4Gb RAM. All CPI systems use local formulae with only one hidden atom to create the initial grounding $\mathbf{G}^0$.

### 5.1 SEMANTIC ROLE LABELLING

Semantic Role Labelling refers to the task of identifying and classifying the arguments and modifiers of verbs in natural language text, as in

[A0He] [AM-MODwould] [A0n't] [Vaccept] [A1anything of value].

for the verb "accept". Labels such as "A0" serve as placeholders for actual roles of the given verb, such as "acceptor" in the above case. The most effective approach to Semantic Role Labelling to date is based on the output of a constituent parser. Each constituent is labelled with the type of argument or modifier it represents with respect to the verb in question. We model the task using a (typed) binary *label* predicate defined over constituents and possible labels. Atoms of this predicate are hidden at test time.

We set up a knowledge base of rules that describe local features of constituents and global dependencies between labels, resembling previous work in Semantic Role Labelling [Punyakanok et al., 2005]. The rules we use are slightly more general than our examples $\phi_1$ and $\phi_2$ in section 2.

We learn the weights of this model using the CoNLL 2005 dataset [Carreras and Marquez, 2005] and the Online Learner MIRA [Crammer and Singer, 2003]. For inference during training we use CPI with ILP.

For testing we use the first 100 verb frames from the WSJ test set of the CoNLL 2005.[7] In table 1 we show the following metrics: 1) the score delta to the optimal solution with respect to the soft clauses, $\Delta s_{soft} = s_{\mathbf{G}_{soft}}(\hat{\mathbf{y}}, \mathbf{x}) - s_{\mathbf{G}_{soft}}(\mathbf{y}', \mathbf{x})$ where $\mathbf{G}_{soft}$ contains all ground formulae for each non-deterministic formula; 2) the number of violations of deterministic formulae; 3) the runtime taken; 4) the number of calls to the optimiser; 5) the F1 accuracy on the task. Note that the total score delta $\Delta s = s(\hat{\mathbf{y}}, \mathbf{x}) - s(\mathbf{y}', \mathbf{x})$ is always dominated by the hard constraints as they have very large weights. Thus if system A produces one less violation than system B its total score delta $\Delta s$ will be smaller.

We first note that using plain MWS with 100,000 flips (M-100k) and no restarts[8] is less accurate in terms of soft model score and F1 accuracy when compared with CPI-MWS using the same number of flips (C-M-100k). It is also significantly slower and produces some hard constraint violations while CPI & MWS does not. When using ILP we achieve perfect model score since ILP returns exact solutions. Using ILP with CPI (C-ILP) is still exact; however, CPI speeds up the solver by almost two orders of magnitude.[9]

We also ran CPI-ILP on the full test set to compare our system with the state of the art, yielding 77.1 F1 measure. When compared to the entries in the CoNLL shared task that only use the output of one parser our system would come out first [Carreras and Marquez, 2005].

|  | $\Delta s_{soft}$ | Viol. | $t$(s) | It. | F1 |
|---|---|---|---|---|---|
| M-100k | -0.098 | 0.05 | 70 | 1 | 0.61 |
| C-M-10k | -0.26 | 0.01 | 0.18 | 3.7 | 0.65 |
| C-M-100k | -0.074 | 0 | 1.2 | 3.5 | 0.69 |
| ILP | 0 | 0 | 4.6 | 1 | 0.79 |
| C-ILP | 0 | 0 | 0.065 | 3.2 | 0.79 |

Table 1: Semantic Role Labelling results averaged over the first 100 examples in WSJ test set; $\Delta s_{soft}$ is the soft score delta wrt to the true MAP solution; $Viol$ the number of violations; $It.$ the number of optimisation calls; $t(s)$ the time spent in seconds; F1 is F1 accuracy.

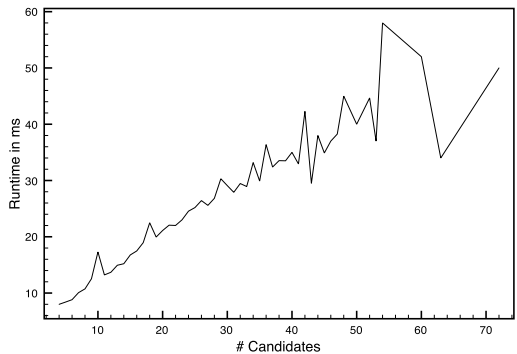

Figure 1: Runtime of CPI-ILP for Semantic Role Labelling for different numbers of candidate constituents.

We also wanted to investigate how CPI scales with problem size. Figure 1 shows the runtime of CPI-ILP against the number of candidate nodes. It seems that for this dataset and problem, CPI scales almost linearly with problem size up until 50 candidates. After 50 candidates a linear trend can only be guessed due to data sparseness. This linear dependency is interesting because the actual number of ground formulae rises quadratically (due to the no-overlap formula and others).

### 5.2 JOINT ENTITY RESOLUTION

Entity resolution is a crucial problem in many business, government and research projects. It can be described as the task of finding database records that refer to the same entity and is very similar to Coreference Resolution in NLP. In our experiments records are citations, and we search for citations describing the same publication; however, we not only want to find matching citations, we also want to jointly identify author or venue name strings referring to the same author or venue, respectively.

We use a knowledge base with 46 formulae provided by Singla and Domingos [2005] with predicates such as *sameBib* and *sameAuthor* that denote citation and author matches, respectively. The knowledge base states regularities such as "if two authors names match the corresponding citations match" or "if the tdf-if distance between the titles is between 0.7 and 0.8 the titles match". It also contains the transitivity rule

$$sameBib(v_1, v_2) \wedge sameBib(v_2, v_3)$$
$$\Rightarrow sameBib(v_1, v_3)$$

---

[7] The reason for not using more instances were memory problems we encountered when we were using MWS alone and grounding the complete network. These problems will likely disappear when using LazySAT instead of MWS.

[8] Note that we also experimented with using restarts; however, differences to runs with equivalent total numbers of flips and no restarts were only marginal.

[9] Note that the difference in runtime between MWS and ILP should be taken with caution: for ILP we use a well-established software library (lp-solve), for MWS our own implementation.

|        | $\Delta s_{soft}$ | Viol. | $t$(m) | It. | F1 |
|--------|-------------------|-------|--------|-----|------|
| M-100k | -4578 | 704.5 | 2.67 | 1 | 0.30 |
| C-M-1k | -2446 | 796.1 | 0.55 | 30 | 0.69 |
| C-M-10k | -2682 | 594.9 | 0.96 | 30 | 0.70 |
| C-ILP | 0 | 0 | 1.56 | 5.9 | 0.72 |

Table 2: Averaged results over 10 folds of the Cora dataset; $\Delta s_{soft}$ is the soft score delta wrt to the true MAP solution; $Viol$ the number of violations; $It.$ the number of optimisation calls; $t$(m) the time spent in minutes; F1 is F1 accuracy.

This is a hard constraint and imposes a difficult problem for many generic inference methods [Poon and Domingos, 2006].

In our first set of experiments we used a cleaned version [Singla and Domingos, 2005] of the Cora Database [Bilenko and Mooney, 2003], containing approximately 1200 citations of computer science articles. In total these citations refer to about 120 unique publications. Following previous work [Singla and Domingos, 2006b], we tested and trained using a 10-fold leave-one-out procedure and Pseudo-Likelihood estimation while ensuring that folds do not contain split citation clusters [Singla and Domingos, 2005]. Each fold contains roughly 120 records.

Table 2 shows our results for Entity Resolution. They are consistent with those in table 1: again CPI renders MWS faster and more accurate both in terms of violations, soft model score and F1 measure. However, this time we cannot directly compare plain ILP with CPI-ILP because the full ILP did not fit into memory. In other words, here CPI makes an infeasible method feasible. Note that in this case CPI-MWS did not converge, thus we terminated the algorithm after a predefined number of iterations (30).

Interestingly, the F1 accuracy of CPI with MWS is significantly better than the F1 accuracy of MWS alone. This can be explained if we consider that CPI-MWS returns solutions with significantly higher soft model score. This score reflects what the model learnt to be a good matching, independent of the number of violations. CPI-MWS can achieve a higher soft score because it starts at a solution that maximises the local score without considering any hard constraints. MWS, on the other hand, starts at a random solution that may contain many (locally) unlikely matchings and will spend most of its time making these unlikely matchings consistent.

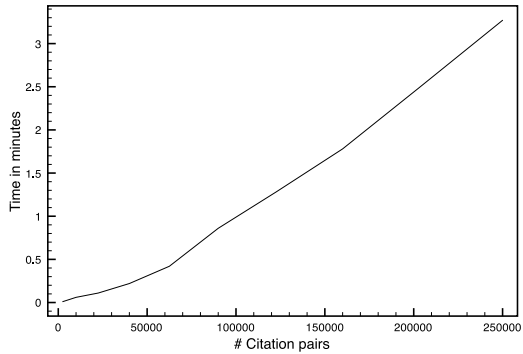

Figure 2: Runtime of CPI-ILP for different subsets of Bibserv.org, averaged over 5 instances for each number of citation pairs.

In our second set of experiments we wanted to again evaluate the runtime behaviour of CPI when the problem size increases. For this purpose we used the Bibserv.org corpus and a model trained on the Cora dataset. Bibserv.org consists of about 20,000 citations. We used the same random subsets of size 50 to 500 in steps of 50 as Singla and Domingos [2006b]. All following results are averaged over 5 datasets of the same size.

Figure 2 shows the runtime of CPI-ILP with increasing number of citation pairs – this corresponds to the number of decisions to make. Again CPI-ILP seems to scale linearly with the number of variables and thus quadratically with the number of citations. Yet, the number of features in the complete network scales at least cubically with the number of citations due to the transitivity clause.

## 6 CONCLUSION

In this paper we presented Cutting Plane Inference (CPI), a novel method for finding MAP solutions in Markov Logic that incrementally solves partial versions of the complete Ground Markov Network based on a Cutting Plane approach. While Cutting Planes have been used for specific MAP inference problems before, this work shows how they can be generalised and incorporated into a Statistical Relational Learning framework where they become automatically available for a wide range of tasks. Our method essentially serves as a *meta* algorithm that alternates between deterministic first order query processing on one hand, and numeric optimisation of partial problems on the other.

We evaluated the proposed algorithm using two real-world tasks for which we showed MWS to perform poorly: Joint Entity Resolution and Semantic Role Labelling. In both cases CPI makes an exact method (ILP) more efficient while remaining exact, and an approximate method (MWS) both faster and more accurate.

However, exact MAP inference in general Graphical Models is PP-complete [Park, 2002]. Thus we obviously cannot expect Cutting Plane Inference to work for arbitrary formulae, weights and problems – at least not for ILP as base solver. Yet, we believe that both of the above tasks are instances of a larger class of problems that are not so much characterised by their network structure or strength of weights but by how well local formulae and weights predict the global goodness of a structure. CPI extends the applicability of SRL to this class and might therefore contribute to a more widespread use of SRL.

It will be important to investigate how to characterise the class of problems we can not solve with CPI. For example, consider a conjunctive formula like $p(v_1) \wedge q(v_2)$ with positive weight and sparsely populated predicates $p$ and $q$ in the current solution $\mathbf{y}$. Separation will find all pairs of $v_1$ and $v_2$ for which the conjunction does not hold and here this set would be almost exhaustive, resulting in a problem not much smaller than the original one.